\pdfoutput=1
\documentclass[11pt]{article}
\usepackage{authblk}
\usepackage{naacl2021}
\usepackage{graphicx}
\usepackage{latexsym}
\usepackage{times}
\usepackage{soul}
\usepackage{url}
\usepackage{times}
\usepackage{graphicx}
\usepackage{subfigure}
\usepackage{amssymb}
\usepackage{soul}
\usepackage{url}
\usepackage{amsmath}
\usepackage{subfigure}
\usepackage{booktabs}   
\usepackage{multirow}
\usepackage{algorithmic}
\usepackage{times}
\usepackage{latexsym}

\usepackage[T1]{fontenc}
\usepackage[utf8]{inputenc}

\usepackage{microtype}

\title{A Question-answering Based Framework for Relation Extraction Validation}
\author[1*]{Jiayang Cheng}
\author[2*$\dagger$]{Haiyun Jiang}
\author[1]{Deqing Yang}
\author[1$\dagger$]{Yanghua Xiao}
\affil[1]{Fudan University, \texttt{\{chengjy17,yangdeqing,shawyh\}@fudan.edu.cn}}
\affil[2]{Tencent AI Lab, \texttt{haiyunjiang@tencent.com}}
\affil[*]{\textit{Equal contribution to this work.}}

\begin{document}
\maketitle
\begin{abstract}

Relation extraction is an important task in knowledge acquisition and text understanding.
Existing works mainly focus on improving relation extraction by extracting effective features or designing reasonable model structures.
However, few works have focused on how to validate and correct the results generated by the existing relation extraction models. 
We argue that validation is an important and promising direction to further improve the performance of relation extraction.
In this paper, we explore the possibility of using question answering as validation.
Specifically, we propose a novel question-answering based framework to validate the results from relation extraction models. 
% need to add some explanation or comments for the framework
Our proposed framework can be easily applied to existing relation classifiers without any additional information. 
%While improving the performance of relation extraction, our framework does not require any extra information, which highlights its practical value. 
We conduct extensive experiments on the popular NYT dataset to evaluate the proposed framework, and observe consistent improvements over five strong baselines.
% Our data and codes will be released on GitHub.
%with CNN, PCNN \cite{Lin2016NeuralRE} and their improved versions \cite{han2018hierarchical} as the base models.
\end{abstract}

\section{Introduction}
Relation extraction (RE) aims to identify the target relations of entity pairs based on their contexts.
It is typically modeled as the relation classification (RC) problem with pre-defined relation classes \cite{Zhou2005ExploringVK,Zeng2015DistantSF,Lin2016NeuralRE,han2018hierarchical,Jiang2019RelationEU}. 
For example, the entity pair (\emph{Microsoft, Bill Gates}) should be classified into the relation \texttt{founder} given the context ``Bill Gates co-founded Microsoft with his childhood friend Paul Allen".

RE has been extensively studied for many years.
The early works mainly build relation extractors using hand-crafted features \cite{Zhou2005ExploringVK} or kernel methods \cite{Bunescu2005ASP}.
In recent years, deep learning-based RE models are extensively proposed and they are powerful to utilize the background knowledge of entities and learn implicit features from complex sentences \cite{Zeng2014RelationCV,Zeng2015DistantSF,pawar2017relation,Feng2018ReinforcementLF}.
In general, most of the existing RE works can be categorized into the \emph{model-level} research, that is, they improve the performance of RE from the perspective of \emph{feature engineering} or \emph{model design}.
%Significant progress has been made \cite{pawar2017relation,Smirnova2018RelationEU} in recent years.
%We categorize these works into model-level research.

However, for multiple reasons, e.g., data noise \cite{Lin2016NeuralRE,Luo2017LearningWN}, limited data size \cite{Han2018FewRelAL} or difficulty in structure selection \cite{Raschka2018ModelEM}, it is increasingly hard to significantly improve the performance through feature engineering or model design. 
%Instead, we argue that \emph{the validation at the result level} is also important to further improve the state-of-the-art RE performance.
In this paper, we argue that \emph{the validation at the result level} is also important to further improve the state-of-the-art RE performance.
That is, given the results predicted by a RE model, we hope to detect the wrong predictions and further correct them. 
%In this way, many mispredicted entity pairs by the RE model are promising to find the correct relations.
%To the best of our knowledge, no previous work has been done in this direction.

%However, it is unrealistic for the RE model to conduct the validation by itself, since there is no information indicating the correctness of the predicted relations.
It is unrealistic for the RE model to conduct the validation by itself, since there is no information indicating the correctness of the predicted relations.
% (usually decided by the predicted confidence score)   ???????????
%model or feature design has an upper limit on improving the performance.
%That is, 
%As a result, it is not easy to construct a high-accuracy relation extraction model.
%That is, the RE models usually make wrong relation predictions for certain entity pairs, and the RE models fail to correct such errors by themselves. 
This necessitates the introduction of an additional model (i.e., validation model) for our purpose. 
%the wrong predictions cannot be easily detected and corrected by the RE models themselves. 
%We hope to design a validation mechanism to check the correctness of the predicted relations, and then correct the wrong results.
Nevertheless, it is difficult to get explicit supervision to learn a validation model that could identify the correctness of a candidate relation.
Fortunately, we observe that relation classification (RC) is closely related to the task of knowledge base completion \cite{Wang2014KnowledgeGE, shi2018open}.
The former aims to predict the semantic relation given the head and tail entities, while the latter tries to predict the tail entity given the head entity and the relation.

Inspired by this intrinsic relatedness, we try to validate the candidate relations by the task of tail entity prediction with the help of question answering (QA) models.
Specifically, given an entity pair and a candidate relation, we construct a question based on the head entity and the relation. %, assuming the tail entity in the context sentence is the true answer of the question. 
Then we predict whether the question can be answered with the tail entity using a QA model.
Intuitively, when the candidate relation is correct, the constructed question is expected to be confidently answered by the QA model with the tail entity (under the context sentence).
But when the candidate relation is incorrect, the QA model should give a low score, because the tail entity is no longer a valid answer.

Our validation process can be summarized as the following four steps.
First, given an entity pair, a relation classifier is learned to generate the score distribution over the pre-defined relations based on the context sentences.
Second, a small set of candidate relations are selected, since it is unnecessary to validate all the relations.
Then, for each selected relation, we employ the validation model to evaluate the correctness of these candidate relations. 
Finally, a more reasonable score for each candidate relation is estimated based on the results from the classifier and the validation model.
A detailed example is presented in Sec 2.2.

% Although generation-then-validation is promising to improve the RE performance, the choice of validation model is challenging, since there is no explicit supervision to train a model that could identify the correctness of a candidate relation.
%Nevertheless, the choice of validation model is challenging, since there is no explicit supervision to train a model that could identify the correctness of a candidate relation.
%%directly indicates the correctness of each candidate relation.
%Fortunately, we observe that relation classification (RC) is closely related to knowledge base completion (tail entity prediction) \cite{shi2018open}.
%The former aims to predict the target relation given the head and tail entities, while the latter tries to predict the tail entity given the head entity and the relation.

% Intuitively, if both the two questions can be confidently answered, which indicates the contextual sentence clearly expresses the candidate relation, then this relation is very likely to be correct.
% Since we construct two questions for each relation in two opposite directions\footnote{That is, the question about the head (tail) entity is constructed to infer the tail (head) one.}, we name our validation model as \emph{bi-directional} question answering (Bi-QA).

% ??????????????sentence??????????context

\emph{Contributions.}
%We summarize our contributions as follows. 
%(1) To the best of our knowledge, we are the first to focus on the validation and correction task in relation extraction.
(1) To the best of our knowledge, we are the first to focus on using question answering-based tail entity prediction to validate the relation extraction results. 
%of the results generated by the existing RC models.
%(2) We propose a novel generation-then-validation framework with QA as validation.
(2) We propose a reasonable QA-based validation framework and prove its effectiveness through extensive experiments.
It is noteworthy that our framework can be used as a plug-and-play component for most existing relation classifiers without additional information, which highlights its practical value.
(3) To improve validation efficiency, two effective relation selection strategies are also proposed.

%We conduct extensive experiments on four strong baselines, and prove the effectiveness of the proposed framework.

\begin{figure*}[!htbp]
	\centering
	\includegraphics[scale=0.72]{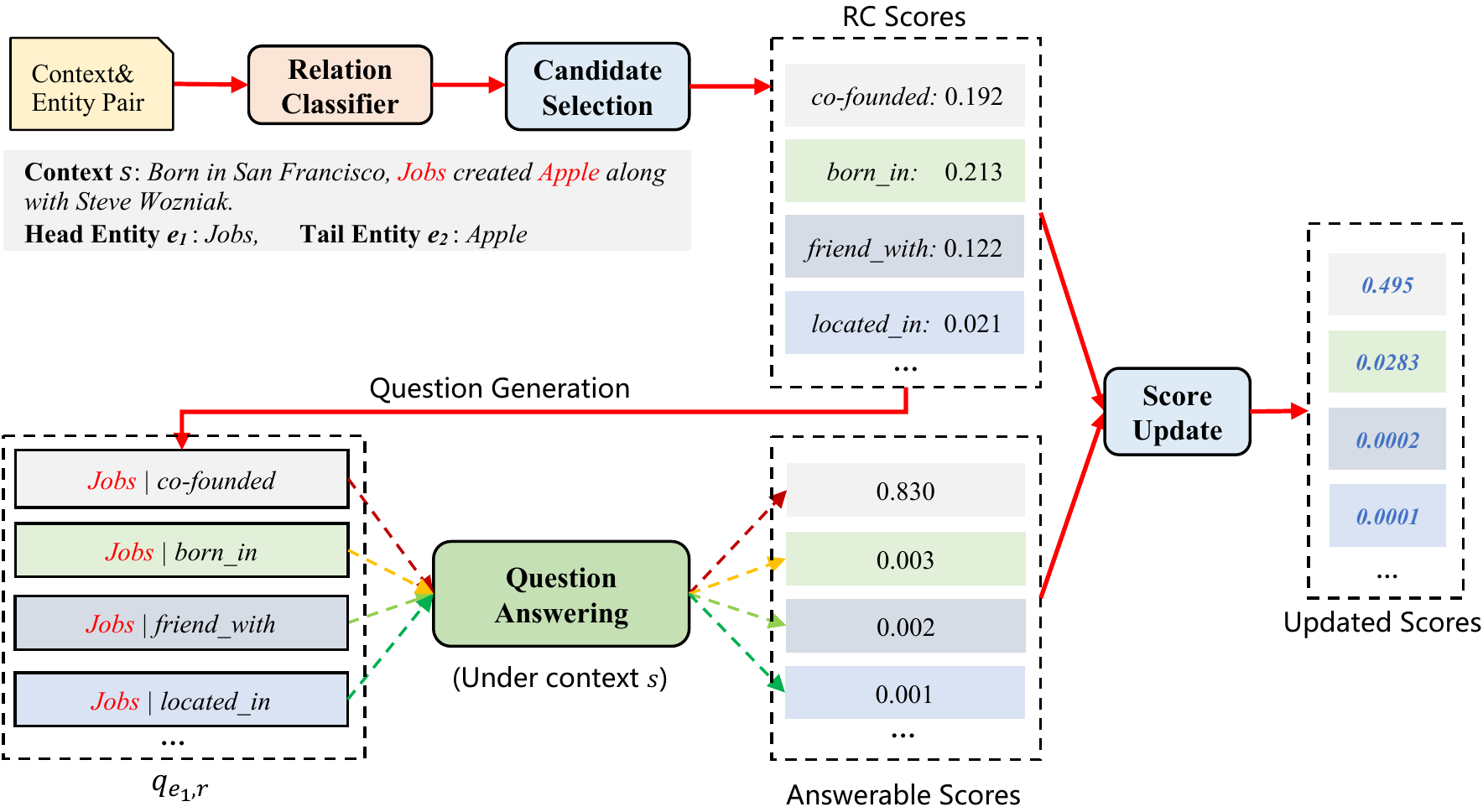}
% 	\caption{An example to illustrate the idea of generation-then-validation. The true relation for (\emph{Jobs, Apple}) is \emph{co-founded}. The RC model is first used to generate the scores for all the relations. Then we select a samll set of promising relations for further validation that is realized by a QA model. Finally, the updated scores are obtained by combining the scores from RC and QA models.} 
	\caption{An example to illustrate the idea of QA based validation. The true relation for (\emph{Jobs, Apple}) is \emph{co-founded}. The RC model is first used to generate the scores for all the relations. Then we select a samll set of promising relations for further validation that is realized by a QA model. Finally, the updated scores are obtained by combining the scores from RC and QA models.} 
	\label{validation_example}
\end{figure*}
%
%\begin{figure}[!htbp]
%	\centering
%	\includegraphics[scale=0.41]{generation_then_validation.pdf}
%	\caption{The overview of the generation-then-validation framework.}
%	
%	\label{generation_then_validation}
%\end{figure}

%\vspace{-0.5cm}
\section{Overview}
\subsection{Framework Overview}
Denote the relation extraction dataset as $\mathcal D = \{(e_1,e_2,s;r)\}$, where $r \in \mathcal{R}$ is the true relation of the entity pair $(e_1,e_2)$ and $\mathcal R = \{r_1,...,r_l\}$ is the pre-defined relation set with $|\mathcal R| = l$.
The input of our framework is an entity pair $(e_1,e_2)$ with context $s$, and the outputs are the scores of all the relations in $\mathcal R$.
In the framework, a relation classifier is first trained to predict the score $p_j$ for each relation $r_j$ (Sec 3.1).
Assume $k$ ($k \ll l$) candidate relations are selected.
For each selected $r_j$, we construct a question $q_{e_1, r_j}$ (Sec 3.2) and employ a QA model to check its correctness (Sec 3.3).
Finally, the updated score for $r_j$ is obtained based on the scores from the relation classifier and the QA model (Sec 3.4).
Besides, the candidate relation selection strategies are also described in Sec 3.4.

%We first predict the score for each relation 
%top-$k$ candidate relations by a classifier for $(e_1,e_2)$. 

%The validation score generated from the question answering model is denoted as $p_{j, QA}$.

%As a result, the relation score distribution is obtained for $(e_1,e_2)$. 
%Finally, the score for each relation is updated based on the classifier and the Bi-QA model.
%??A global picture of the structure is presented in figure \ref{generation_then_validation}.

\subsection{Example}
% We illustrate the generation-then-validation process in Figure \ref{validation_example}, which mainly contains four steps:
We illustrate the pipeline in Figure \ref{validation_example}, which mainly contains four steps:
% \paragraph{(1)  \emph{Generation.}}
\paragraph{(1)  \emph{RC score generation.}}
Given $(e_1,e_2)$ = (\emph{Jobs, Apple}) and its context $s$, a relation classifier first generates the scores for all the pre-defined relations.

\paragraph{(2) \emph{Candidate relation selection for validation.}}
Intuitively, it is costly to conduct the validation for all the relations.
Moreover, this is also unnecessary, since most relations can be filtered out by the RE model.
Alternatively, we select a small set of candidate relations for further validation.
A direct strategy is to chose the top-$k$ relations with the highest scores (predicted by the RE model) as candidates,  e.g., \emph{co-founded} and \emph{born\_in} in Figure \ref{validation_example}.
This is reasonable since most wrong relations are predicted with lower scores by the RE model and we only need to focus on the relations with higher scores.
Besides, this paper also presents another effective strategy to select the promising relation subset (see Sec 3.4).

\paragraph{(3)  \emph{Question construction and validation.}}
For each selected candidate relation $r_j$, we validate the correctness with the help of a QA model.
We construct the question $q_{e_1,r_j}$ by directly concatenating the head entity $e_1$ with the candidate relation $r_j$.
For example, when $e_1$, $r_j$ and $e_2$ are \emph{Jobs},  \emph{co-founded} and \emph{Apple}, question $q_{e_1,r_j}$ is {``\emph{Jobs} $|$ co-founded''}.

Then, given the context $s$, a QA model is employed to judge whether $q_{e_1,r_j}$ can be answered with the tail entity $e_2$.
Intuitively, the context sentence $s$ clearly expresses the relation \emph{co-founded}, thus the question ``\emph{Jobs} $|$ co-founded'' should be confidently answered with the tail entity \emph{Apple}. Therefore, in Figure \ref{validation_example}, we can observe a high QA score (0.83).
Meanwhile, the sentence does not express all the other candidate relations, so the questions for these relations should be scored lower. %, i.e., the estimated scores are lower. 
%For instance, question ``\emph{Jobs} $|$ born in'' cannot be answered with \emph{Apple}, so it will be given a lower score (0.003).

%In practice, the validation score is defined as a combination of the answerable probability and the confidence probability, which we will explain in the next section.

\paragraph{(4) \emph{Score update.}}
For each candidate relation $r_j$, we obtained two scores from the RC and QA models, respectively.
The updated score for $r_j$ is a combination of the two scores and the details are shown in Sec 3.4.
In general, the updated score is more reasonable than the individual score from either the RC or the QA model.
%Since there are in total $|\mathcal{R}|$ candidate relations for each item, it is sometimes unnecessary to validate all of them, as the relation classification model has been proved to be able to filter out some wrong candidates. Based on this intuition, we experiment on two kinds of combination of the scores, and get similar improvement on performance.

%, i.e., the estimated scores are lower. 
%For instance, question ``\emph{Jobs} $|$ born in'' cannot be answered with \emph{Apple}, so it will be given a lower score (0.003).

%In practice, the validation score is defined as a combination of the answerable probability and the confidence probability, which we will explain in the next section.

%Since there are in total $|\mathcal{R}|$ candidate relations for each item, it is sometimes unnecessary to validate all of them, as the relation classification model has been proved to be able to filter out some wrong candidates. Based on this intuition, we experiment on two kinds of combination of the scores, and get similar improvement on performance.

%\section{Generation-then-validation Framework}
\section{QA-based Validation Framework}

\subsection{Relation Classification (RC)}
In general, the task of RE is popularly modeled as the problem of relation classification (RC).
%the existing RC models can be classified into two categories: sentence level and entity pair level.
%Sentence level RC models use one context sentence to predict the target relation, while the entity pair level models use a bag of sentences.
%Thus, from the perspective of input setting, the former is the special case of the latter, i,e., the sentence bag only contains one sentence.
%In this paper, we focus on the entity pair level RC.
%Then, the context $s$ denotes a bag of sentences.
Given the context $s$ of an entity pair $t$, the context representation $\boldsymbol s$ is learned through an encoder, e.g., a convolutional network \cite{Zeng2015DistantSF}, a recurrent network \cite{Zhang2015RelationCV}, or a self-attention network \cite{Du2018MultiLevelSS}.
During this process, some extra information may also be incorporated into the encoder to improve the performance. 
For instance, position embeddings \cite{Zeng2014RelationCV} and domain knowledge \cite{Zhou2005ExploringVK,Weston2013ConnectingLA,Li2019ChineseRE}.
The representation $\boldsymbol s$ is further used to predict the score distribution over the pre-defined relation set $\mathcal R$, which is denoted as $\{p(r_1|\boldsymbol s),...,p(r_l|\boldsymbol s)\}$.

% of the bag of sentences is obtained using strategies like attention mechanism \cite{Lin2016NeuralRE,han2018hierarchical}.
%Finally, $\boldsymbol s$ is used to predict the score distribution over the pre-defined relation set $\mathcal{R} = \{r_1,...,r_l\}$, which is denoted as 
%The general objective of the relation classification model can be denoted as $\mathcal L_{RC}({\Theta})$, where $\Theta$ are the parameters to be learned.

\subsection{Question Construction}
Given an entity pair $t=(e_1,e_2)$ and a candidate relation $r_j$, we construct a question $q_{e_1,r_j}$ for $e_1$ and $r_j$.
%we consider an incomplete triple $(e_1,r_j,*)$ and check whether the missing entity $e_2$ can be correctly inferred from the context $s$.
%We construct a question $q_{e_1,r_j}$ for the triple, and use a QA model to predict the probability that $e_2$ is the answer under the context $s$.
In this paper, the question is constructed by directly concatenating the head entity string $e_1$ with the relation string $r_j$. 
%Intuitively, $q_{e_1,r_j}$ is meaningful only when $(e_1,r_j,*)$ is a \emph{valid} triple, i.e., the type of $e_1$ is consistent with that of the subject of $r_j$.
%Otherwise, the question will be meaningless.
For example, in Figure \ref{validation_example}, $(\emph{Jobs, co-founded})$ will generate the question ``\emph{Jobs} $|$ co-founded''.
Intuitively, $q_{e_1,r_j}$ is meaningful only when $e_1$ matches $r_j$ in semantics, otherwise, $q_{e_1,r_j}$ is meaningless (e.g., ``\emph{Jobs} $|$ located in''), which indicates $r_j$ is not the correct relation.
Thus, the question itself provides some informative features to help us identify the correctness of the relation.

\subsection{Question Answering Model}
%While improving the performance of relation extraction, our framework does not require any extra information, which highlights its practical value.

% notations and structures
\subsubsection{Dataset generation}
Instead of using additional information, we train our validate (QA) model by constructing samples based on the RE dataset, which highlights the practical value of the framework since additional information cannot be easily obtained.
Given an entity pair $t = (e_1, e_2)$ and its context $s$ in the RE dataset $\mathcal{D}$, for each candidate relation $r_j \in \mathcal{R}$, a question $q_{e_1, r_j}$ is first constructed. 
We also add an additional token ``null'' in the first position of $s$.
Then if $r_j$ is the correct relation of $t$, the answer of this question is $a_{e_1, r_j} = e_2$, and the answerable flag $f_{e_1, r_j}$ is set as 1 (\textit{True}); otherwise, $a_{e_1, r_j}$ is set as ``null'' (the first token in $s$) and $f_{e_1, r_j}$ is 0 (\textit{False}).
Then, the QA dataset $\mathcal{Q}$ can be denoted as $\{  (q_{e_1, r_j}, s; a_{e_1, r_j}, f_{e_1, r_j})  \}$. %where $q_{e_1, r_j}$ and $a_{e_1, r_j}$ are the question and answer, $f_{e_1, r_j}$ is the flag indicating whether the constructed question can be answered by $e_2$.

%Given the question-answer set $\mathcal{Q}$, the question answering model is trained or fine-tuned on the set $\mathcal{Q}_{train}$; then it is tested on $\mathcal{Q}_{test}$ to generate the validation scores $\{p_{j, QA}\}$.

%(similarity of our task and tasks in squad v1.0 and squad v2.0)
\subsubsection{Model selection}

In this paper, we fine-tune the pre-trained ALBERT \cite{lan2019albert} as our QA model to fulfill the validation task.
 %a QA model is employed to check whether the tail entity $e_2$ is a valid answer for the generated question $q_{e_1, r_j}$ under the context $s$. 
%In this paper, we consider ALBERT \cite{lan2019albert} to fulfill the validation task. 
As a lite version of BERT \cite{devlin2019bert}, ALBERT significantly reduces the size of parameters as well as the training time. 
Meanwhile, it achieves new state-of-the-art results on the SQuAD 2.0 \cite{rajpurkar2018know} and other NLP datasets. 
%With such advantage, it is then practical to employ it to validate the relation classification results.

%This is desirable, because intuitively, models that can distinguish the answerable questions from the unanswerable ones should also be able to discern the nuances between our wrongly constructed questions and the correct ones. Third, many work have been done on this dataset\footnote{https://rajpurkar.github.io/SQuAD-explorer/}. At the time of writing, the top model has achieved an EM of 90.002 and an F1 of 92.425, both higher than the human performance. This provides us with many choices.

%(our choice (also mention that the choice could be flexible))

%In our generation-then-validation framework, validation is after generating relation extraction scores. Therefore, we hope the validation model scales well so that the framework is feasible in practice. 
%Given this, we choose 

\subsubsection{Fine-tuning details}
Based on the existing work \cite{lan2019albert}, we fine-tune the pre-trained ALBERT model on our dataset $\mathcal{Q}$. 
Our QA model will predict two results for each sample: the start and end positions of the answer in the context and the answerable probability.
Specifically, given a sample $\psi = (q_{e_1, r_j}, s; a_{e_1, r_j} f_{e_1, r_j})$ in $\mathcal{Q}$, we formalize the answer prediction as follows. 
For the $l$-th position (corresponding to word $w_l$) in the context $s$, the QA model predicts two probabilities: $p_l^{st}$ and $p_l^{ed}$, where $p_l^{st}$ ($p_l^{ed}$) denotes the probability that $w_l$ is the starting (ending) word of the answer. 
Besides, the model also predicts an answerable probability $p_{ans}$. 
Thus, the loss functions for each sample $\psi$ are defined as follows:
\begin{equation} \label{eqn2}
\small
\begin{split}
&loss_{position}^{\psi} = -\log p_{l_{s}}^{st} -\log p_{l_{e}}^{ed},  \\
&loss_{ans}^{\psi} = -f_{e_1, r_j} \log p_{ans} - (1-f_{e_1, r_j}) \log (1-p_{ans}),
\end{split}
\end{equation}
where $f_{e_1, r_j}$ is the answerable flag in sample $\psi$.
$l_{s}$ ($l_{e}$) denotes the true start (end) position of the answer $a_{e_1, r_j}$ in $s$. 
%That is, $t_{st}$ and $t_{ed}$ are set as 0 (``null'' as answer) if $q_{e_1, r_j}$ cannot be answered from the context $s$.
If $a_{e_1, r_j}=$ ``null'' (i.e., unanswerable), then $l_{s}$ and $l_{e}$ are set as 0.
The loss $loss_{position}^{\psi}$ is about the position prediction while $loss_{ans}^{\psi} $ is about the answerable prediction.
%the question $q_{e_1, r_j}$ cannot be answered from the context $s$, then $l_{s}$ and $l_{e}$ are set as 0, i.e., taking the token ``null'' as the fake answer. 

%ground-truth label (1 for answerable and 0 for unanswerable) of the answerable flag in sample $\psi$.

The loss function on $\mathcal{Q}$ is defined as:
\begin{equation}
\small
\mathcal L_{QA}({\Phi}) = 
\frac{1}{2N} \sum_{\psi \in \mathcal Q}  \left({loss}_{position}^{\psi} + loss_{ans}^{\psi} \right),
\label{loss fucntion for QA}
\end{equation}
where $N=|\mathcal{Q}|$ and $\Phi$ are the parameters to be fine-tuned.

\subsubsection{Validation score generation}
The validation (QA) model can be well fine-tuned by minimizing $\mathcal L_{QA}({\Phi})$ in Equation \ref{loss fucntion for QA}.
During the test phase, we first construct the sample $\psi = (q_{e_1, r_j}, s; a_{e_1, r_j}, f_{e_1, r_j})$ for the candidate relation $r_j$ (to be validated) given an entity pair $t$ and $s$.
Then the answerable, start and end probabilities are computed by the QA model. 
The validation score is defined as follows:
\begin{equation}
\small
\begin{split}
&p_{j, QA} = \sqrt{p_{ans} \times p_{confidence}} ,\\
&p_{confidence} =   
\max_{\mbox{\tiny$\begin{array}{c} 
		i,j\\ 
		i,j\neq 0, i\leq j \end{array}$}}    p_{i}^{st} p_{j}^{ed},
\end{split}
\label{validation socre}
\end{equation}
%where $p_{ans} = 1 - p_{ans}$.
%\begin{equation}
%\small
%p_{confidence} =  
%\max_{\mbox{\tiny$\begin{array}{c} 
%		i,j\\ 
%		i,j\neq 0, i\leq j \end{array}$}}    p_{i}^{st} p_{j}^{ed},
%\end{equation}
where $i$ ($j$) denotes the $i$-th ($j$-th) position in $s$.
That is, $p_{confidence}$ is the maximum probability among all the candidate answer strings (except ``null'') in $s$.

$p_{confidence}$ measures the confidence level of the answerable score $p_{ans}$.
That is, if $q_{e_1, r_j}$ is answerable, then the QA model will give a very high score on a string within $s$.
%if the QA model gives a very high score on a string within $s$, we consider $q_{e_1, r_j}$ tends to be answerable.
In contrast, if the question is unanswerable, any string in $s$ cannot be the answer and $p_{confidence}$ will be very small.

%We conclude that, apart from the answerable probability, we also make use of the maximum probabilities of the possible answer strings to measure the correctness of $r_kj$. 
%
%
%The definition of $p_{position}$ is motivated by the following observation.
% 
%%\sqrt{\max\limits_{k} p_k^{st} \times \min\limits_{l} p_l^{ed}}$ (wrong for the confidence). 
%
%% (illustrate why we combine the above stuffs together like that)
%
%The definition of  is intuitive: if the QA model gives a very high score on a string within the sentence, we consider this question tends to be answerable, and vice versa. 
%%Combining this measurement of confidence with the answerable probability gives the final validation score.  

\subsection{Candidate Relation Selection and Score Update}
Given a test entity pair and its context, we select a small set of promising relations for validation.
In this paper, we provide two effective relation selection strategies.

%Given an entity pair and its context, for each candidate relation, the validations score can be obtained based on Equation \ref{validation socre}.
%Then, a natural question is, which relations should be selected for validation for each entity pair?
%Here we introduce and discuss about two validation strategies.

\subsubsection{Strategy I}
The first strategy is to select the most confidently predicted scores by the QA model for validation. 
%The basic assumption is that the highly confident scores by the QA model tend to be reliable.
Specifically, for each test entity pair and all the relations in $\mathcal{R}$, we generate the corresponding QA scores and then sort them in descending order.
Intuitively, extreme scores, either very high or very low, indicate that the QA model confidently expresses whether a question is answerable or not.
%the very high (low) scores indicate the QA model confidently believes the corresponding question is answerable (unanswerable).
However, the scores in the middle range indicate less confidence.
So we retrieve the top $\alpha$ and last $\beta$ percent ones among all the scores for validation.
If the relation $r_j$ is retrieved, its final score is updated by 
\begin{equation}
\small
p'_{j}  = \left [(p_{j, QA})^{\lambda} \times p_{j}\right ]^{\frac 1 {\lambda+1}}.
\label{scoreupdate}
\end{equation}

%In this way, we obtain $|\mathcal{R}|\times|\mathcal{D}_t|$ scores and we sort them in descending order.

As shown in Equation \ref{scoreupdate}, the final score $p'_j$ for $r_j$ is computed by a weighted multiplication of its QA score $p_{j, QA}$ and the RC score $p_j$, where $\lambda >0$ is a predefined constant that determines the relative importance between the two scores.

Otherwise, if a relation is not selected for validation, its updated score is computed by:
\begin{equation}
\small
p'_{j}  = \left [c^{\lambda} \times p_{j} \right ]^{\frac 1 {\lambda + 1}},
\label{non k scoreupdate}
\end{equation}
where $c\in (0, 1)$ is a constant.
%Explain why there is a factor of (the power of one of them), different in scale. 

%In this way, the scores for all the candidate relations are updated for a given entity pair.

%As mentioned in introduction, we can select the top-$k$ relations with the highest scores by the RC model.
%Besides, we also introduce another validation strategy in this section.
%Next, we describe the two strategies in detail.

%In this subsection, we discuss the two score update strategies we use in our work. 
%Essentially, they are from different perspectives. 
%The first update strategy tends to trust the scores from the relation extraction model, and uses the scores from the question answering model as an auxiliary validation source. 
%In contrast, the second update strategy tends to trust the scores from the question answering model. 
%It primarily considers the question answering scores, and search from the question answering scores to decide which scores to update.

\subsubsection{Strategy II}
Given an entity pair, strategy II aims to validate the top-$k$ relations based on the scores from the RC model, which has been illustrated in Figure \ref{validation_example}.
The basic assumption is that the relations predicted by the RC model with higher scores tend to contain the correct relation.

To be specific, for each entity pair we choose $k$ (out of $|\mathcal{R}|$) relations with the highest scores from the RC model. 
Then, we compute the corresponding validation scores using the QA model, and use them to update the scores of the top-$k$ candidate relations.
If the relation $r_j$ is in the top-$k$ set, then the score is updated according to Equation \ref{scoreupdate}. 
For the rest relations (out of top-$k$) in $\mathcal{R}$, their updated scores are obtained by Equation \ref{non k scoreupdate}.

\paragraph{Discussion}
Essentially, strategy I and II are designed from different perspectives. 
Strategy I tends to trust the scores from the validation (QA) model while strategy II tends to trust the results from the RC model.
In general, strategy I is more suitable for the scene where the validation model is reliable and the pre-defined relation set is small.
This is because strategy I needs to compute the QA scores for all the relations given an entity pair.
In contrast, strategy II is more suitable for the scene where the RC model is reliable, because the selected relations are decided by the RC model.
Besides, strategy II is not sensitive to the size of the relation set, since it does not need to compute all
 the QA scores of relations.

\section{Experiments}
\subsection{Experimental Details}

\subsubsection{Dataset description}
We evaluate our proposed framework on the NYT dataset \cite{Riedel2010ModelingRA}, which is widely used in the field of RE. 
%Using Freebase as its knowledge base, the NYT dataset is constructed by extracting relations from the New York Times corpus.
The dataset contains 53 relations, including a special relation ``NA" that indicates there is no pre-defined relation between the given entity pair.
There are 522,611 sentences, 281,270 entity pairs and 18,252 relation facts in the training set.
The testing set contains 172,448 sentences, 96,678 entity pairs and 1,950 relation facts.

%Table \ref{statisticsofnyt} gives several statistics of the training and testing sets.

%\begin{table}
%	\centering
%	\caption{Several statistics of the NYT dataset}
%	\label{statisticsofnyt}
%	\begin{tabular}{lllll}
%		\toprule
%		& Sentences & Entity pairs & Relational facts\\
%		\midrule
%		Training & 522,611 & 281,270 & 18,252 \\
%		Testing & 172,448 & 96,678 & 1,950\\
%		\bottomrule
%	\end{tabular}
%\end{table}

%\subsubsection{Baselines}
%% baseline as in the relation classifiers cnn+att, pcnn+att, ...
%Given the nature of our generation-then-validation framework, our baselines are the aforementioned relation extraction models, i.e. NRE (CNN/PCNN+ATT) and HNRE (CNN/PCNN+HATT). They are trained and tested on the NYT dataset, and then used for comparison with their validated version.

\subsubsection{Relation classifiers}
In our experiments, we choose five relation classifiers as baselines and try to improve their results using our framework.
\paragraph{(1) CNN+ATT and (2) PCNN+ATT \cite{Lin2016NeuralRE}.}
CNN+ATT takes a convolutional neural network (CNN) to extract informative features from sentences.
Besides, sentence-level attention (ATT) is used to alleviate the noise in a sentence bag.
In PCNN+ATT, the CNN encoder is replaced with the piecewise convolutional neural network (PCNN), where the piecewise max pooling operation is adopted for feature extraction.
For CNN/PCNN+ATT, we use the open-source toolkit\footnote{https://github.com/thunlp/OpenNRE}.

\paragraph{(3) CNN+HATT and (4) PCNN+HATT \cite{han2018hierarchical}.}
Based on CNN/PCNN, \cite{han2018hierarchical} proposes a hierarchical attention mechanism that exploits hierarchical information of relations, thus generating CNN+HATT and PCNN+HATT.
The hierarchical attention has been proven to be very effective and CNN/PCNN+HATT obained state-of-the-art performance on the NYT dataset.
For CNN/PCNN+HATT, we use the official code\footnote{https://github.com/thunlp/HNRE}.

\paragraph{(5) RESIDE \cite{reside2018}.}
In RESIDE, the relation alias information and entity types are introduced into RE task as soft constraints for relation prediction.
To integrate these information,  RESIDE employs graph neural networks (GNNs) to encode syntactic information from text.
For RESIDE, we also use the official code\footnote{https://github.com/malllabiisc/RESIDE}.

\subsubsection{Experimental setup}
% parameter settings (some description, and a table)
%For the score generation, we train the NRE (CNN/PCNN+ATT) and HNRE (CNN/PCNN+HATT) models on the NYT dataset. 

%For convenience, they are shown in table \ref{scoregenerationpara}.
%%After training, we test the RE models on the testing set and get scores for each item.
%
%\begin{table}
%	\centering
%	\caption{Parameter settings for the score generation models, where ``$\sim$'' denotes ``CNN/PCNN''}
%	\label{scoregenerationpara}
%	\begin{tabular}{llll}
%		\toprule
%		& $\sim$+ATT  & $\sim$+HATT \\
%		\midrule
%		Batch size & 160 & 160 \\
%		Learning rate & 0.01 & 0.2 \\
%		Word dimension &  50 & 50 \\
%		Position dimension & 5 & 5 \\
%		Drop-out probability & 0.5 & 0.5 \\
%		Sentence embedding size & 230 & 230 \\
%		Convolution kernel size & 3 & 3 \\
%		\bottomrule
%	\end{tabular}
%\end{table}

% params cnn+att, pcnn+att, albert, combination params
\begin{table}
	\centering
	\caption{Parameter settings for fine-tuning the ALBERT-base model.}
	\label{albertpara}
	\scalebox{0.8}{
		\begin{tabular}{lll}
			\toprule
			Batch size & 16 \\
			Learning rate & $1.5\times 10^{-5}$ \\
			Max sequence length & 384 \\
			Ratio of positive to negative samples & 1:2 \\	 % ?????????????????????????????????????????????????????????
			Optimizer &Adam \\
			\bottomrule
		\end{tabular}
	}
\end{table}

% ??????????????????????????????????????????
% PR????????????
\subsection{Results and Analysis}
%PR-AUC, 
\begin{table}
	\centering
	\caption{Area Under precision/recall Curves for the two validation strategies. We conclude that \emph{CNN+HATT+ ValStrgy I improves  the AUC of 4\% compared with CNN+HATT}.}
	\label{auc}
	\scalebox{0.8}{
		\begin{tabular}{cc|cc}
			\toprule
			\multicolumn{4}{c}{AUC(\%)} \\
			\midrule
			CNN+ATT & 32.81  & PCNN+ATT & 34.66 \\
			+ValStrgy I & 34.42 & +ValStrgy I &  36.28 \\
			+ValStrgy II &  33.99  & +ValStrgy II &  35.55 \\
			
			\midrule
			CNN+HATT &  41.75  & PCNN+HATT &  41.97 \\
			+ValStrgy I & 43.54 & +ValStrgy I & \textbf{43.62}\\
			+ValStrgy II & 43.11  & +ValStrgy II & 43.26\\
			\midrule  
			RESIDE &  0.386       \\
			+ValStrgy I & 0.399   \\
			+ValStrgy II & 0.401   \\
			
			\bottomrule
		\end{tabular}
	}
\end{table}

\begin{table}
	\centering
	\caption{Parameter settings for the two strategies.}
	\label{conspara}
	\scalebox{0.8}{
		\begin{tabular}{cccccccc}
			\toprule
			Parameters&$k$ &$c$ &$\lambda$ &$\alpha $&$\beta$ \\
			\hline
			Strategy I &- &0.9&10&10&20 \\
			Strategy II &3 &0.9&10&-&- \\
			\bottomrule
		\end{tabular}
	}
\end{table}

\begin{figure*}[!htbp]
	\centering
	\subfigure[]{
		\includegraphics[width=6cm]{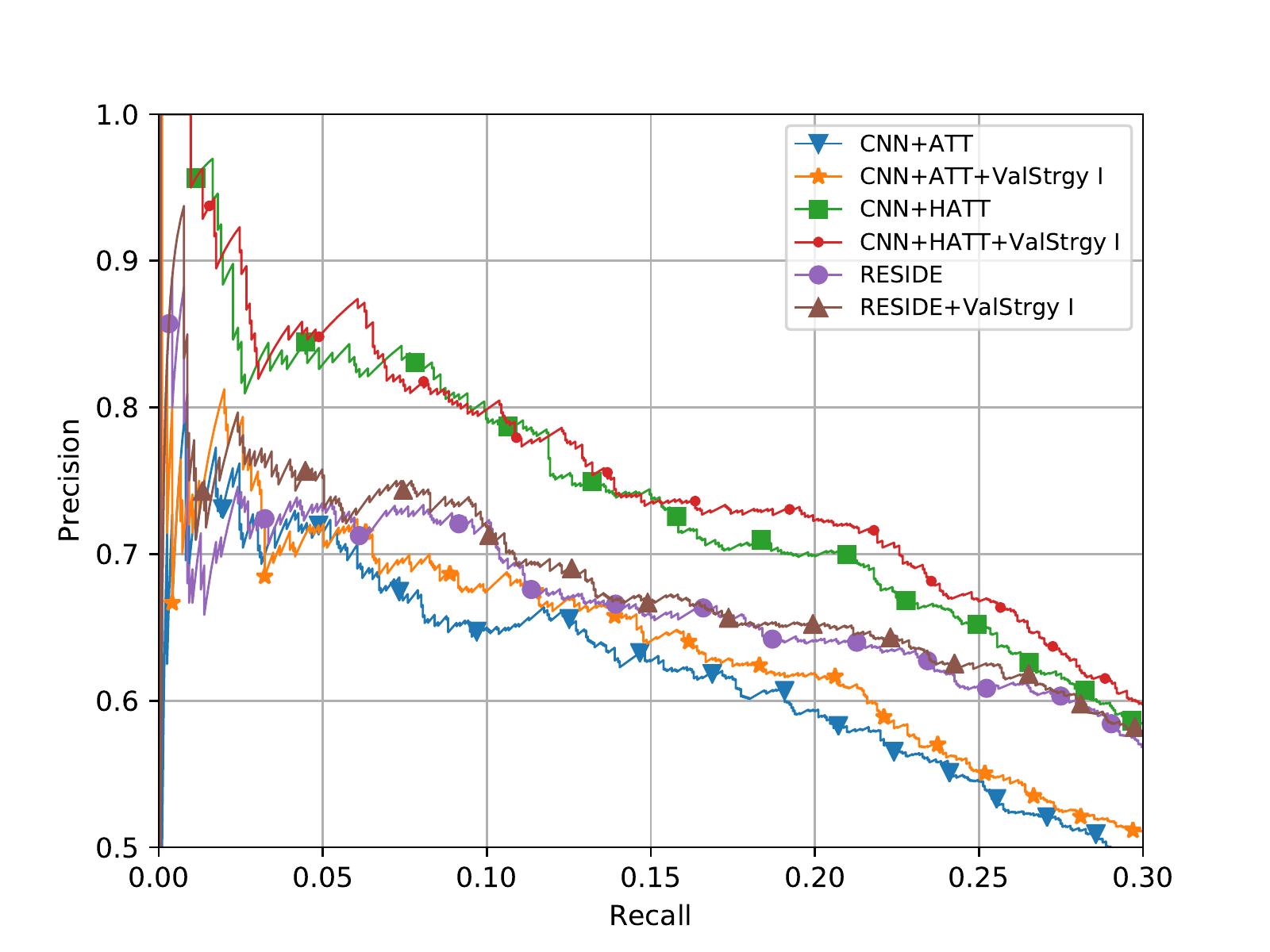}
		%\caption{fig1}
	}
	\quad
	\subfigure[]{
		\includegraphics[width=6cm]{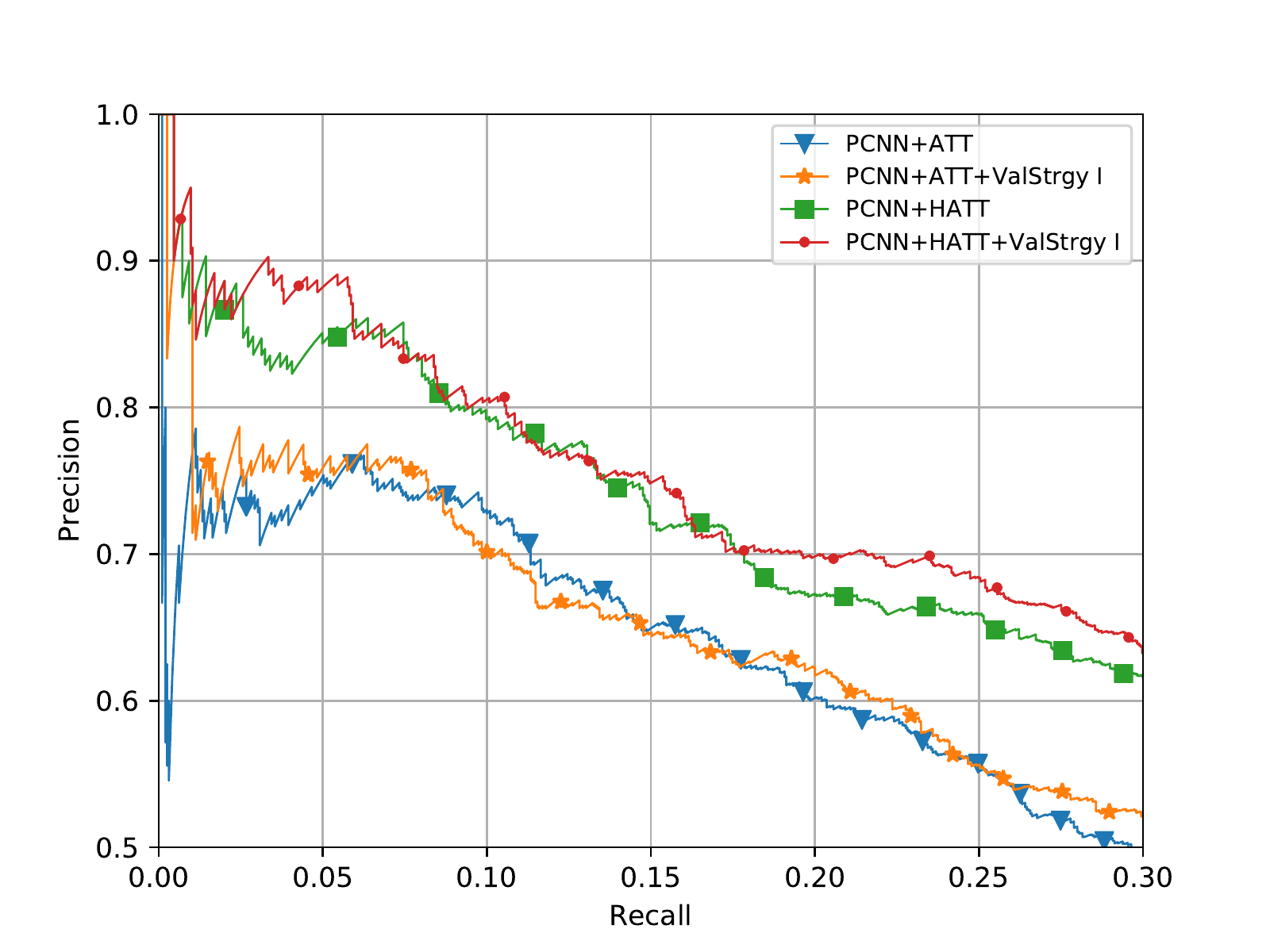}
	}
	\quad
	\subfigure[]{
		\includegraphics[width=6cm]{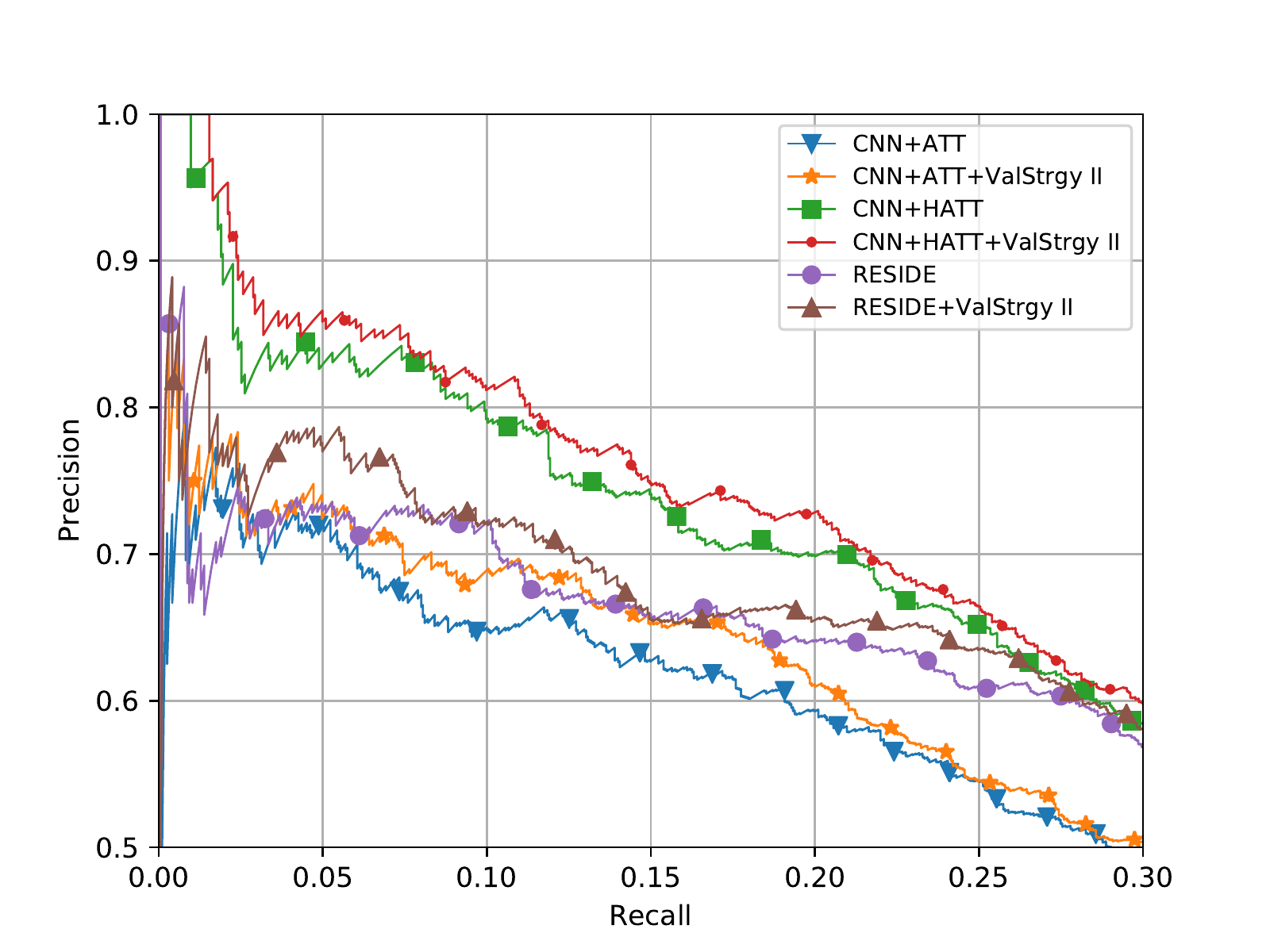}
	}
	\quad
	\subfigure[]{
		\includegraphics[width=6cm]{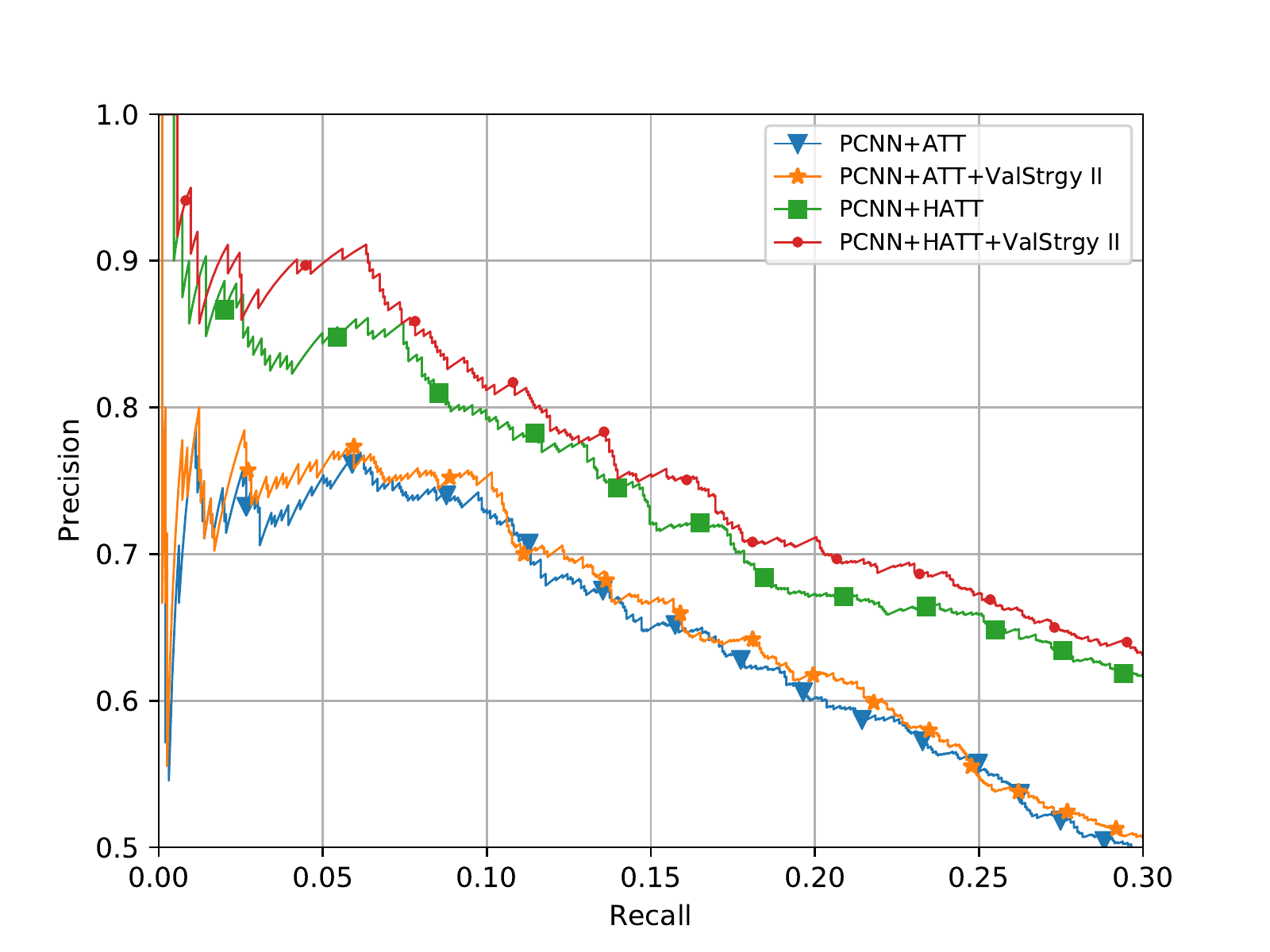}
	}
	
	\caption{(a) and (b): Aggregate precision/recall curves under strategy I; (c)  and (d): Aggregate precision/recall curves under strategy II.}
	\label{prcurve}
\end{figure*}

% finetune albert model 
%On training the ALBERT model, we fine-tune the pretrained ALBERT-base model, as described in Section 2.4. 
For the relation classifiers, we follow the default parameter settings as described in the original papers.
The details of our validation (QA) model has been described in Sec 3.3, and we implement it based on the open source\footnote{https://github.com/kamalkraj/ALBERT-TF2.0}.

Note that many entity pairs in NYT contain multiple sentences.
To construct samples for these entity pairs, we concatenate the multiple sentences to generate the context $s$.
To control the length of the resulting context, and to reduce the noise brought by the parts of sentence far from the head and tail entities, we cut each sentence by the position of its head and tail entities. 
Specifically, the substring between the 40 tokens before the head entity and 40 words after the tail entity is preserved, and other parts are abandoned. 
Empirically, the cut sentences are shorter than 100 words, and thus the length of the concatenated context is acceptable. 
%Together, these sentences can express the target relation. 
%However, the QA model only takes one context for each question. 
%To resolve this conflict, we concatenate the multiple sentences to generate a context. 

%%%???????????????????????paper?????????????????????????na??????????????????????????????????????????????????��??????????????????????????????????????????????????????????????????????????

%As described in the previous section, we construct question-answer pairs from the original NYT dataset for training and testing. 
%For convenience, the resulting sets are of the same format as the SQuAD 2.0 dataset. 

We also explain how to deal with the special relation ``NA"  during validation.
Since ``NA'' has no specific semantics, we did not construct samples involving ``NA'' when training our QA model.
As a result, the QA score for ``NA'' will not be generated during validation.
That is, in both the two relation selection strategies, the relation ``NA'' will be ignored although it is selected and the score of ``NA'' will be updated using Equation \ref{non k scoreupdate}.

In our validation (QA)  model, the ratio of positive (answerable) to negative (unanswerable) samples is set as 1:2.
Other parameter settings in the QA model are shown in Table \ref{albertpara}.
Besides, the parameters used in our two strategies (Section 3.4) are presented in Table \ref{conspara}. 
All the parameters are obtained using cross-validation by splitting a subset from the training dataset.
% ???????????????????????????????????????????????????????1:2, ????????????
%In the NYT dataset, there are many items labeled as ``NA", which means the distant supervision method decides there is no relation between the head and tail entities. 
%We do not consider these items. Instead, for each non-NA item, we randomly sample negative relations from the other candidate relations to generate the negative samples for training set.

% ????????title??????????????????????????????; ????????????????????????????????????????????????????????????????????????????????????????????????????????????????????????????????????????????; ?????????????????????????????????????????????????????????????????????????????0???????????????
%The questions and answers are constructed following the description in the previous section. For positive items, the answer is the tail entity, and the ``is\_impossible" flag is set to be \textit{false}. For negative items, the answer is set empty, and the ``is\_impossible" flag is set to be \textit{true}.

% ?????????????????????generation ???? ALBERT ??????????model????????????????????????????????????????????????????????????????
%Similarly, the testing set is constructed from the NYT dataset, except that we do not conduct the negative sampling, i.e., all items are positive samples in the testing set. After fine-tuning the ALBERT model on the training set, we test it on the testing set to get the question answering scores. 

% ????????????????????????????????????????????????????????????????????????????

% Precision@N
\begin{table*}
	\centering
	\caption{Precision@N for the two validating strategies.}
	\label{patn}
	\scalebox{0.8}{
		\begin{tabular}{lcccc|lcccc}
			\toprule
			Precision@$N$(\%) & $N$=100 & $N$=200 & $N$=300 & Mean  & & $N$=100 & $N$=200 & $N$=300 & Mean \\
			\midrule
			CNN+ATT & 71.0 & 67.5 & 65.0 & 67.8  & PCNN+ATT & 73.0 & 74.0 & 71.3 & 72.8  \\
			+ValStrgy I & 70.0 & 69.0 & 68.7 & 69.2 & +ValStrgy I & 84.0 & 82.0 & 77.0 & 81.0 \\
			+ValStrgy II & 72.0 & 71.0 & 69.3 & 70.8 & +ValStrgy II & 75.0 & 75.5 & 70.7 & 73.7 \\
			
			\midrule
			CNN+HATT & 84.0 & 82.0 & 77.0 & 81.0 & PCNN+HATT & 83.0 & 81.5 & 77.3 & 80.6\\
			+ValStrgy I & 85.0 & 81.0 & 78.3 & 81.4 & +ValStrgy I & 89.0 & 81.5 & 78.0 & 82.8\\
			+ValStrgy II & 85.0 & 82.5 & 78.3 & 81.9  & +ValStrgy II & \textbf{90.0} & \textbf{83.5} & \textbf{78.7} & \textbf{84.1} \\
						\midrule
			RESIDE &   72.0&73.0&69.0&71.3& \\
			+ValStrgy I & 76.0&75.0&70.0&73.7& \\
			+ValStrgy II & 78.0&74.0&72.0&74.7&   \\
			
			\bottomrule
		\end{tabular}
	}
\end{table*}

% ???????????????????????????????????????????????????????????????????????????????????????????????????????????????????????
In this section, we present the results and analysis on the proposed framework. 
Specifically, we compute the aggregate precision/recall curves, the Area Under precision/recall Curves (AUC) and the Precision@$N$ for the five classifiers as well as their validated versions (both by strategy I and strategy II).
Precision@$N$ denotes the precision of the top $N$ predicted relational fact.
The results updated by strategy I (strategy II) is denoted as ``+ValStrgy I" (``+ValStrgy II").

\subsubsection{Overall evaluation results}
Table \ref{auc} gives the AUC results, which quantify the overall performance of the models. 
We empirically compare the results from our two validation strategies with the baseline models. 
From Table \ref{auc} we observe that: 
\begin{itemize}
	\item  (1) After applying the QA task as validation (with both the two strategies), the performance of all the models can be effectively improved. 
%	It indicates that our generation-then-validation framework is effective for all the baselines.
	It indicates that our QA-based validation framework is effective for all the baselines.
	This is because some wrong predictions are corrected during the validation process.

	\item  (2) CNN/PCNN+ATT/HATT use CNN or its variant as the sentence encoder while RESIDE takes GNN to learn the features from sentences.
	Both CNNs and GNNs are two representative neural network structures.
	They learn the relation-aware features in sentences from different perspectives.
	Using our framework, the performance of all the baselines are successfully improved, which indicates the validation model can learn complementary features that are not captured by both the CNN/GNN-based classifiers.
	
	\item  (3) In general, the improvement after applying validation strategy I is more significant than strategy II. 
	It indicates that the top $\alpha$ and last $\beta$ percent scores are more reliable for validation in our experiments.
	%One of the reasons is that ALBERT is a very powerful pre-trained model 
	In particular, \emph{PCNN+HATT+ValStrgy I obtains new state-of-the-art results on the NYT dataset.}
	Also note that, though strategy II is slightly inferior to strategy I, it takes much less time, as it does not require to compute all the relations in $\mathcal{R}$ for each entity pair in advance. 
	
\end{itemize}

Figure \ref{prcurve} shows the aggregate precision/recall curves. 
For clarity, we present the results from each validation strategy in two figures.
Subplot (a) and (b) give the results from validation strategy I; (c) and (d) give the results from validation strategy II.
We have the following observations from Figure \ref{prcurve}:
(1) Generally, curves from both validating strategies are on top of the ones of the baselines. This means both validating strategies are effective in improving the performance.
(2) On subplot (b) and (d), for baselines PCNN+ATT/HATT, when recall is within the interval $(0, 0.1)$, there are significant improvements in precision for both strategies. It indicates that the validations successfully filter out the wrong high score predictions by validating with QA scores.

\begin{table*}
	\centering
	\caption{The details of the two examples used in the case study.}
	\label{case}
	\scalebox{0.78}{
		\begin{tabular}{c|c|c|p{11cm}}
			\toprule
			Example & {Entity Pair} & True relation   & \qquad  \qquad  \qquad  \qquad \qquad\qquad Context \\
			\midrule
			\#1 & (Cook county, Chicago) & \emph{contains}   & Now, the state is retooling the program to include all of \textbf{Cook county}, which encompasses \textbf{Chicago} and many of its suburbs .\\
			\midrule
			\#2 & (Powerset, San Francisco) & \emph{place\_founded} & ... a friend suggested he check out a \textbf{San Francisco} start-up, \textbf{Powerset}, which is trying to build a rival search engine.\\
			\bottomrule
		\end{tabular}
	}
\end{table*}

\subsubsection{Effect of the validation on the high score predictions}
We present the Precision@$N$ results for the two validating strategies in Table \ref{patn}. 
The Precision@$N$ results show how the validation affects the precision of the top $N$ predictions. 
From Table \ref{patn} we observe that:
(1) After the validation, there are improvements in the mean values of Precision@$N$ in all five baselines with two validating strategies. 
This means the QA model manages to discern the wrong high score predictions.
(2) After the validation, improvements on PCNN-based models are much more significant than on the CNN based models. 
This coincides with our observation on the precision/recall curves. %???????????????????????????????????
(3) We also observe that the Precision@$N$ results from strategy II are better than these from strategy I, which is different from the conclusion of the AUC metric.
This phenomenon indicates strategy II generates higher scores for the correct relations compared with strategy I.
But strategy I is better at reducing the score bias for more relations, which generates better overall performance (AUC).

%Our experiments find that most of top $k=3$ relations predicted by the RC model

%One of the reasons is that strategy II is able to strengthen the scores of the top $k=3$ relations predicted by the RC model and most of them are corrected, which makes the Precision@$N$ results higher. 

\subsection{Case Study}

\begin{table}
	\centering
	\caption{The detailed scores of the two examples predicted by our framework, where the updated scores are obtained using Equation \ref{scoreupdate} with $\lambda=10$.}
	\label{caseScore}
	\scalebox{0.8}{
		\begin{tabular}{c|cc}
			\toprule
			Example  \#1& Relation 1 (True)& Relation 2 (False) \\
			\toprule
			    & \emph{contains} & \emph{neighborhood\_of} \\
			
			 RC score & 0.1330 &  0.2221 \\
			 QA score & 	0.9997	  & 	0.0057	\\
			 Updated score & 0.8322 & 0.0080 \\
			 
			 \midrule
			\midrule
		Example  \#2& Relation 1 (True)& Relation 2 (False) \\
			 \midrule
			  & \emph{place\_founded} & \emph{place\_lived} \\
			
			 RC score & 0.0031 & 0.0533 \\
			 QA score &  0.8990	   & 	0.0073	\\
			 Updated score & 0.5369 & 0.0087\\
			\bottomrule
		\end{tabular}
	}
\end{table}

% In this section, we give two examples showing how our generation-then-validation framework works.
In this section, we give two examples showing how our QA based validation framework works.
Due to limited space, we only consider one classifier (PCNN+ATT) in our case study\footnote{Similarly, other RC models can also be analyzed.}. 
In Table \ref{case} we present the basic information of the two examples.
The corresponding RC scores, QA scores and the updated scores for the correct and wrong relations are presented in Table \ref{caseScore}. 
%The results are from PCNN+ATT, and the parameter $\lambda$ in score update is $10$.

In Example \#1, the true relation between \emph{Cook county} and \emph{Chicago} is \textit{contains}.
However, the RC model PCNN+ATT wrongly predicts the relation \textit{neighborhood\_of} as the target relation, i.e., outputting the score of 0.2221 for \textit{neighborhood\_of} and only 0.133 for \textit{contains}.
However, in the QA model, the question constructed by \emph{neighborhood\_of} cannot be answered by the tail entity \emph{Chicago}. 
As a result, the QA model gives a very low score to the relation \emph{neighborhood\_of}. 
Instead, it gives a high score to \textit{contains}. 
After updating the original RC score with the QA score, the score of the true relation \textit{contains} increases to 0.8322, while the score of \textit{neighborhood\_of} decreases to 0.0080.

\section{Related Work}
The early work for relation extraction manually designs a variety of relation-aware features \cite{Zhang2006ACK}.
%However, feature engineering is noticeably expensive, which makes it difficult to apply to large-scale relation extraction.
In recent years, deep neural networks have been extensively used in RE task \cite{Zeng2015DistantSF,Du2018MultiLevelSS,Han2018FewRelAL,Zhang2018GraphCO,Guo2019AttentionGG,Li2019EntityRelationEA}.
%Deep learning is very powerful for learning implicit features for relation prediction.
%For example, \cite{Zeng2015DistantSF} combine multi-instance learning (MIL) with piecewise convolutional neural networks to choose the most informative sentence for RE.
%\cite{Zhang2018GraphCO} combines dependency parse features with GNNs to considerably increase the performance of RE systems. 
%For example, \cite{Zeng2015DistantSF, Zhang2018GraphCO} 
%\cite{Han2018FewRelAL} proposes a deep prototype network method to few-shot relation classification.

Many references also combine QA with information extraction \cite{Jijkoun2004InformationEF,Yao2014InformationEO,Qiu2018QA4IEAQ,Li2019EntityRelationEA}.
%\cite{Jijkoun2004InformationEF} investigates the impact of the precision/recall trade-off of information extraction on the performance of an offline corpus-based QA system.
\cite{Yao2014InformationEO} shows that, with the help of information extraction, the QA task over structured data outperforms most baselines.
\cite{Levy2017ZeroShotRE} models RE as a simple QA problem, i.e., giving the head entity and the relation and predicting the head entity.
\cite{Qiu2018QA4IEAQ} builds a model to produce high-quality relation triples from sentences by QA.
%The settings of these two methods are similar to our validation model.
Our framework is essentially different from the existing works \cite{Yao2014InformationEO,Qiu2018QA4IEAQ,Li2019EntityRelationEA} that either use QA models to extract triples or take information extraction to assist the QA tasks.
In contrast, we focus on improving the RE task by validation and correctness using QA models.
Moreover, our framework can be applied to any existing RC model.

\section{Conclusion and Future Work}
In this paper, we focus on improving the performance of RE by conducting the validation and correctness of the existing RC models.
The QA task is introduced as the validation task for RC.
% Further, we design a novel generation-then-validation framework that can be applied to any existing relation classifier.
Further, we design a novel QA based validation framework that can be applied to any existing relation classifier.
%we introduce the bi-directional question answering as a validation task to improve the performance of relation extraction.
%To further improve the performance of both the classifier and the Bi-QA model, we introduce the consistency assumption between them and propose a multi-task training strategy with the assumption as posterior regularization.
%model it as a posterior regularization for the multi-task learning.
Besides, we also propose two candidate relation selection strategies to update the relation scores.

We argue that, in addition to the task of RE, \emph{our framework can also be applied to the task of knowledge graph completion}, where a RC model is used as the validation model to check the correctness of the results by the KBC models.
%Although we focus on the task of relation extraction, 
Besides, we will also apply the framework to more information extraction tasks, e.g., entity typing \cite{Choi2018UltraFineET} and slot filling \cite{Zhang2019JointSF}.
%We hope some key challenges in knowledge extraction can be significantly overcome with the help of our framework.

\clearpage
\newpage

\bibliography{anthology,cites}
\bibliographystyle{acl_natbib}

%\appendix
%
%\section{Example Appendix}
%\label{sec:appendix}
%
%This is an appendix.

\end{document}